\documentclass[fleqn,10pt]{wlscirep}
\usepackage[utf8]{inputenc}
\usepackage[T1]{fontenc}
\usepackage{subcaption}
\usepackage{multirow}

\title{Concept Inconsistency in Dermoscopic Concept Bottleneck Models: A Rough-Set Analysis of the Derm7pt Dataset}

\author[1]{Gonzalo N\'apoles}
\author[2]{Isel Grau}
\author[3*]{Yamisleydi Salgueiro}

\affil[1]{Department of Intelligent Systems, Tilburg University, 5037 AB Tilburg, The Netherlands; g.r.napoles@tilburguniversity.edu}
\affil[2]{Information Systems Group, Eindhoven University of Technology, 5600 MB Eindhoven, The Netherlands; i.d.c.grau.garcia@tue.nl}
\affil[3]{Department of Industrial Engineering, Faculty of Engineering, Universidad de Talca, Campus Curic\'o, Curic\'o 3340000, Chile; ysalgueiro@utalca.cl}

\begin{abstract}
Concept Bottleneck Models (CBMs) route predictions exclusively through a clinically grounded concept layer, binding interpretability to concept-label consistency. When a dataset contains concept-level inconsistencies, identical concept profiles mapped to conflicting diagnosis labels create an unresolvable bottleneck that imposes a hard ceiling on achievable accuracy. In this paper, we apply rough set theory to the Derm7pt dermoscopy benchmark and characterize, for the first time, the full extent and clinical structure of this inconsistency. Among 305 unique concept profiles formed by the 7 dermoscopic criteria of the 7-point melanoma checklist, 50 (16.4\%) are inconsistent, spanning 306 images (30.3\% of the dataset). This yields a theoretical accuracy ceiling of 92.1\%, independent of backbone architecture or training strategy for CBMs that exclusively operate with hard concepts. In addition, we characterize the conflict-severity distribution, identify the clinical features most responsible for boundary ambiguity, and evaluate two filtering strategies with quantified effects on dataset composition and CBM interpretability. Symmetric removal of all boundary-region images yields Derm7pt+, a fully consistent benchmark subset of 705 images with perfect quality of classification and no hard accuracy ceiling. Building on this filtered dataset, we present a hard CBM evaluated across 19 backbone architectures from the EfficientNet, DenseNet, ResNet, and Wide ResNet families. Under symmetric filtering, explored for completeness, EfficientNet-B5 achieves the best label F1 score (0.85) and label accuracy (0.90) on the held-out test set, with a concept accuracy of 0.70. Under asymmetric filtering, EfficientNet-B7 leads across all four metrics, reaching a label F1 score of 0.82 and concept accuracy of 0.70. These results establish reproducible baselines for concept-consistent CBM evaluation on dermoscopic data.
\end{abstract}

\begin{document}
	
\flushbottom
\maketitle
\thispagestyle{empty}

\section{Introduction}

Concept Bottleneck Models (CBMs) were introduced by Koh et al.~\cite{koh2020cbm} as a class of two-stage neural architectures that route all predictions through an intermediate layer of human-interpretable concepts before producing a final label. In their original formulation, the backbone first maps the raw input to a vector of concept predictions, and a downstream predictor maps that concept vector to the output class. This design admits two principal variants: soft CBMs, in which the label predictor receives continuous probability scores over each concept, and hard CBMs, in which those scores are discretized into binary or categorical assignments before being passed downstream~\cite{havasi2022leakage}. While soft CBMs achieve higher task accuracy, they suffer from \textit{concept leakage}, a phenomenon in which the label predictor implicitly encodes information beyond the intended concept values~\cite{mahinpei2021promises}. Hard CBMs are resilient to leakage by construction since the bottleneck is a stop-gradient discrete tensor, so the label predictor can only access information that a clinician would observe by reading the concept assignments. Both variants support test-time human interventions, where an expert corrects a mispredicted concept, and the updated prediction propagates forward through the label head, a capability that has been shown to improve accuracy substantially~\cite{shin2023closer}. Extensions of the original framework include Concept Embedding Models~\cite{zarlenga2022cem}, which go beyond the accuracy-vs-interpretability trade-off by learning high-dimensional concept representations. Similarly, Post-hoc Concept Bottleneck Models~\cite{yuksekgonul2023posthoc} retrofit the bottleneck onto already-trained networks by transferring concepts from other datasets without requiring concept-annotated training data. More recent variants, such as Label-free CBMs~\cite{oikarinen2023labelfree} or the UCBM architecture \cite{schrodi2025cbm}, automate concept set construction using vision-language models or latent concept discovery, respectively. These architectures are part of a broader effort to reconcile the accuracy of deep neural networks with the transparency demands of symbolic and logic-based reasoning~\cite{ciravegna2023logic}.

Concept-based models have attracted attention in medical image analysis, where clinicians reason in terms of structured, domain-specific attributes. In dermatology, convolutional neural networks trained end-to-end have achieved accuracy comparable to board-certified dermatologists on skin lesion classification~\cite{esteva2017dermatologist}, yet their black-box nature limits clinical adoption. A systematic review of XAI in skin cancer recognition ~\cite{hauser2022review} found that the overwhelming majority of published studies rely on post-hoc saliency methods, with inherently interpretable approaches remaining substantially underrepresented. Post-hoc methods are widely used because they require no expert-annotated concept data and are readily available in standard deep learning libraries. However, they are frequently rejected as solutions to the transparency problem in clinical settings due to the risk of introducing confirmation bias~\cite{hauser2022review}. CBMs address this gap by grounding predictions in dermoscopic criteria that dermatologists use in practice, such as pigment network morphology, vascular patterns, and the presence of a blue-whitish veil. Lucieri et al.~\cite{lucieri2020interpretability} showed that deep classifiers trained on dermoscopic images encode disease-related concepts in their latent space in a manner consistent with dermatological descriptions. Recent work has extended this to inherently interpretable frameworks, showing that CBMs trained on curated concept annotations can be applied to skin lesion diagnosis without sacrificing classification performance~\cite{patricio2023coherent}. Moreover, concept-grounded approaches yield explanations that post-hoc saliency methods cannot guarantee~\cite{patricio2025twostep}. When concept-grounded XAI explanations are provided alongside predictions, clinicians' diagnostic accuracy, confidence, and trust in the system increase compared to support from a standard non-explainable AI~\cite{chanda2025eyetracking}.

In terms of data, several dermoscopic image datasets have been released to support automated skin lesion classification. The PH$^2$ dataset~\cite{mendonca2013ph2} provides 200 annotated images of melanocytic lesions with expert segmentations, clinical diagnoses, and dermoscopic criteria. HAM10000~\cite{tschandl2018ham} supplies 10,015 multi-source dermatoscopic images across seven diagnostic categories and has become the most widely used benchmark for pigmented lesion classifiers. The BCN20000 dataset~\cite{hernandez2024bcn20000} extends the diversity of available data with approximately 19,000 dermoscopic images from Hospital Cl\'{\i}nic Barcelona, covering challenging lesions such as those in nail and mucosal locations, and was a primary contributor to the ISIC 2019 challenge training set~\cite{combalia2019bcn20000}. The ISIC 2020 challenge introduced a patient-centric collection of over 33,000 images~\cite{rotemberg2021isic2020} that explicitly encodes multiple lesions per patient to support the ugly-duckling diagnostic heuristic. More recently, DERM12345~\cite{yilmaz2024derm12345} released 12,345 images spanning 40 fine-grained subclasses, providing a detailed taxonomic coverage currently available in the open literature. Among all these resources, the 7-Point Checklist dataset, Derm7pt~\cite{kawahara2019derm7pt}, occupies a unique position. Its 1,011 paired clinical and dermoscopic images are annotated not only with diagnosis labels but also with the seven dermoscopic criteria of the 7-point melanoma checklist. It allows for more interpretable model predictions through the inclusion of intermediate semantic attributes~\cite{saeed2025multimodal}. This makes Derm7pt the natural choice for developing and evaluating concept-bottleneck classifiers in dermoscopy, as highlighted in recent concept-based work~\cite{chanda2025eyetracking}. However, this dataset is far from perfect.

In this paper, we propose a formal analysis of concept-level inconsistency in the Derm7pt dataset and its consequences for CBMs trained on dermoscopic data. Using rough set theory~\cite{pawlak1982roughsets, pawlak1991roughsets}, we characterize the full extent of the inconsistency. Among 305 unique concept profiles formed by the seven dermoscopic criteria of the 7-point checklist, 50 (16.4\%) are inconsistent, spanning 306 images (30.3\% of the dataset). This analysis yields a provable theoretical accuracy ceiling of 92.1\% for any classifier depending exclusively on hard concept representations, regardless of backbone architecture, training procedure, or regularization strategy. The bound is tight and holds for any pure CBM variant. A particularly concerning finding is that 136 of the 252 melanoma images (54.0\%) reside in the boundary region, far exceeding the overall boundary fraction of 30.3\%. The ramification of this fact is that the seven checklist concepts are less discriminative for melanoma than for non-melanoma lesions. The concepts most responsible for this boundary ambiguity are irregular dots and globules and irregular streaks, which appear in 79\% and 39\% of boundary-region images compared to 29\% and 19\% of consistent-profile images. To correct the inconsistency, we propose and discuss two filtering strategies. The symmetric strategy removes all boundary images to produce Derm7pt+, a fully consistent subset of 705 images with no hard accuracy ceiling. The asymmetric strategy retains all melanoma images from the boundary while removing only the conflicting non-melanoma samples, yielding 841 images and preserving sensitivity at the cost of partial residual inconsistency. Towards the end of our study, we evaluate a hard CBM across 19 backbone architectures from the EfficientNet, DenseNet, ResNet, and Wide ResNet families.

The remainder of this paper is organized as follows. Section~2 presents the theoretical and statistical analysis of concept inconsistency in Derm7pt using rough set theory, including the derivation of the accuracy ceiling theorem and the characterization of the boundary region. Section~3 describes and evaluates two filtering strategies to remove inconsistencies, resulting in the construction of the cleaned Derm7pt+ dataset\footnote{The resulting Derm7pt+ dataset and the code supporting the experiments of this paper are publicly available at \url{https://github.com/gnapoles/Consistent-Derm7pt}}. Section~4 details the hard CBM architecture used for numerical validation, the data preparation pipeline, and the empirical evaluation across 19 CNN-based backbone architectures. Section~5 concludes with a summary of findings, limitations, and directions for future work.

\section{Dataset Analysis}
\label{sec:analysis}

This section conducts a rough set analysis of the Derm7pt dataset to characterize concept-level inconsistency and its implications for hard CBM architectures. First, we derive a provable accuracy ceiling for any classifier operating exclusively on hard concepts and the whole dataset. Second, we quantify the extent of inconsistency across profiles and images, and identify the dermoscopic criteria most responsible for boundary ambiguity.

\subsection{Theoretical analysis}

Let $U$ be the universe of discourse for the Derm7pt dataset, which contains $|U|=1,011$ dermoscopy images. Each image $x \in U$ is described by a set of clinical concepts $C = \{a_1, \ldots, a_7\}$, where $a_i$ denotes one of the seven dermoscopic criteria of the 7-point melanoma checklist: \textit{pigment network, streaks, pigmentation, regression structures, dots and globules, blue-whitish veil, and vascular structures}. In its binary form, each image also carries a binary diagnosis label $d(x) \in \{0, 1\}$, where 0 encodes non-melanoma, and 1 encodes melanoma. The tuple $S = (U, C \cup \{d\})$ is an information system in the sense of rough set theory. The indiscernibility relation induced by $C$ creates a partition of $U$ into equivalence classes of dermoscopic images that are identical with respect to all concepts. Such an indiscernibility relation is given by:
\begin{equation}
    \mathrm{IND}(C)
    = \bigl\{(x,y) \in U \times U : \forall\, a_i \in C,\;
    a_i(x) = a_i(y)\bigr\}.
    \label{eq:ind}
\end{equation}

The quotient set $E = U / \mathrm{IND}(C)$ is the partition of $U$ into equivalence classes. In this regard, Derm7pt yields $|E| = 305$ distinct concept profiles. Each class $[x]_C \in E$ constitutes a concept profile: the complete signature of dermoscopic features observable through the concept bottleneck. Two images belonging to the same profile are indistinguishable by any concept-only classifier, such as pure bottleneck models that operate with hard concept representations.
	
A concept profile is consistent if all images it contains share the same diagnosis label, and inconsistent otherwise. The positive region of decision $d$ with respect to $C$ is the union of all consistent profiles:
\begin{equation}
    \mathrm{POS}_C(d)
    = \bigcup_{\substack{[x]_C \in E \\
            \exists\, \ell \in \{0,1\}:\, [x]_C \subseteq d^{-1}(\ell)}}
    [x]_C.
    \label{eq:pos}
\end{equation}

\noindent such that $d^{-1}(\ell) = \{x \in U : d(x) = \ell\}$ is the pre-image of label $\ell \in \{0,1\}$. In a binary setting, the boundary region for a given decision class collects all images whose profile contains both diagnosis values:
\begin{equation}
    \mathrm{BND}_C(d) = U \setminus \mathrm{POS}_C(d).
    \label{eq:bnd}
\end{equation}

The quality of classification, which is an RST metric, is defined as follows:
\begin{equation}
    \gamma(C,d) = \frac{|\mathrm{POS}_C(d)|}{|U|}.
    \label{eq:gamma}
\end{equation}

It measures the fraction of images that can be correctly and unambiguously classified by a concept-only rule. A value $\gamma < 1$ signals that the dataset is inconsistent, meaning that no concept-based classifier can perfectly separate the two classes regardless of architecture, training procedure, or supervision scheme.

Let $\Pi$ denote the set of inconsistent concept profiles. For profile $e_k \in \Pi$, let $n_k^{\mathrm{mel}}$ and $n_k^{\mathrm{nmel}}$ be the melanoma and non-melanoma image counts, respectively, with total size $n_k = n_k^{\mathrm{mel}} + n_k^{\mathrm{nmel}}$. The conflict ratio is defined as follows:
\begin{equation}
    \gamma_k
    = \frac{\min\!\bigl(n_k^{\mathrm{mel}},\, n_k^{\mathrm{nmel}}\bigr)}{n_k},
    \qquad \gamma_k \in (0,\, 0.5].
    \label{eq:cr}
\end{equation}
	
A value $\gamma_k = 0.5$ indicates a perfectly ambiguous profile, while $\gamma_k \to 0$ indicates a near-consistent profile where a single annotation outlier creates the conflict. Let $n_k^{\mathrm{maj}} = \max(n_k^{\mathrm{mel}}, n_k^{\mathrm{nmel}})$ be the majority-class count in profile $k$. Therefore, the concept-based accuracy ceiling is given by the following expression:
\begin{equation}
    \mathrm{acc}^{\ast}
    = \frac{|\mathrm{POS}_C(d)| +
        \displaystyle\sum_{k \in \Pi} n_k^{\mathrm{maj}}}
    {|U|}.
    \label{eq:ceil}
\end{equation}
	
\textit{\textbf{Theorem (Accuracy Ceiling)}.}
\textit{For the Derm7pt dataset and the binary settings explained above, any classifier $h: U \to \{0, 1\}$ that exclusively only on the dermoscopic concepts in $C$ (i.e., $h(x) = h(y)$ whenever $(x,y) \in \mathrm{IND}(C)$) satisfies:
\begin{equation}
    \mathrm{acc}(h) \le 0.921
    \quad (92.1\%).
    \label{eq:thm}
\end{equation}
In addition, this performance bound is tight.}
	
\textit{\textbf{Proof.}} For every class $e \subseteq \mathrm{POS}_C(d)$, all images in $e$ share the same label. Because $h$ is constant on $e$, it assigns the correct label to every member of $e$, giving exactly $|\mathrm{POS}_C(d)| = 705$ correct predictions on the positive region.

For each inconsistent concept profile $e_k \in \Pi$, the classifier $h$ must assign a single label to all images in $e_k$, so it correctly classifies at most $n_k^{\mathrm{maj}}$ of them. Summing over all 50 inconsistent concept profiles and using the explicit enumeration of the Derm7pt boundary region, we get the following expression:
\begin{equation}
    |\{x \in \mathrm{BND}_C(d) : h(x) = d(x)\}|
    \le \sum_{k \in \Pi} n_k^{\mathrm{maj}} = 226.
    \label{eq:bnd_bound}
\end{equation}
	
Adding both regions: $|\{x \in U : h(x) = d(x)\}| \le 705 + 226 = 931$, and dividing by $|U| = 1011$ gives the stated bound~(\ref{eq:thm}). Tightness follows from the majority-vote classifier, which assigns the unique label on each consistent profile and the majority label on each inconsistent profile, achieving exactly 931 correct predictions and attaining the bound.

\textit{\textbf{Corollary}. No Concept Bottleneck Model using hard inputs, regardless of training procedure, concept supervision, or regularization, can exceed $92.1\%$ accuracy on the unfiltered Derm7pt dataset for melanoma/non-melanoma classification, because its final decision layer is a deterministic function of the concept vector alone.}

\textit{\textbf{Remark}. The theoretical accuracy ceiling is a tight upper bound that applies to any hard CBM determining class labels exclusively from the seven dermoscopic concepts. This bound is computed globally, treating the entire dataset as a single pool before any partitioning is performed. The local ceiling will vary across training, validation, and test splits depending on how inconsistent concept profiles are distributed among them. A split concentrating boundary-region images in the training set imposes a tighter local ceiling on training accuracy while leaving the test ceiling relatively relaxed.}

\subsection{Statistical Analysis}
	
Applying the rough-set formalization to the Derm7pt dataset for melanoma/non-melanoma classification yields insightful statistics. Of the 305 unique concept profiles, 50 are inconsistent ($|\Pi|=50$, 16.4\%), and 306 images reside in $\mathrm{BND}_C(d)$ (30.3\%). The quality of classification is $\gamma(C,d) = 705/1011 \approx 0.697$, and the accuracy ceiling evaluates to $\mathrm{acc}^{\ast} = 931/1011 \approx 0.921$, as established in the theorem above. This exceeds the majority-class baseline of 75.1\% by 17 percentage points, yet constitutes a non-trivial upper limit. Any CBM reporting accuracy above 92.1\% on the unfiltered dataset is exploiting information that leaks outside the concept bottleneck. A notable finding is that 136 of the 252 melanoma images (54.0\%) fall in $\mathrm{BND}_C(d)$, far exceeding the overall boundary fraction of 30.3\%. This disproportionate representation indicates that the seven checklist concepts in $C$ are systematically less discriminative for melanoma than for non-melanoma lesions.
	
\begin{figure}[!ht]
    \centering
    
    \begin{subfigure}{0.50\textwidth}
    \centering
    \includegraphics[width=\textwidth]{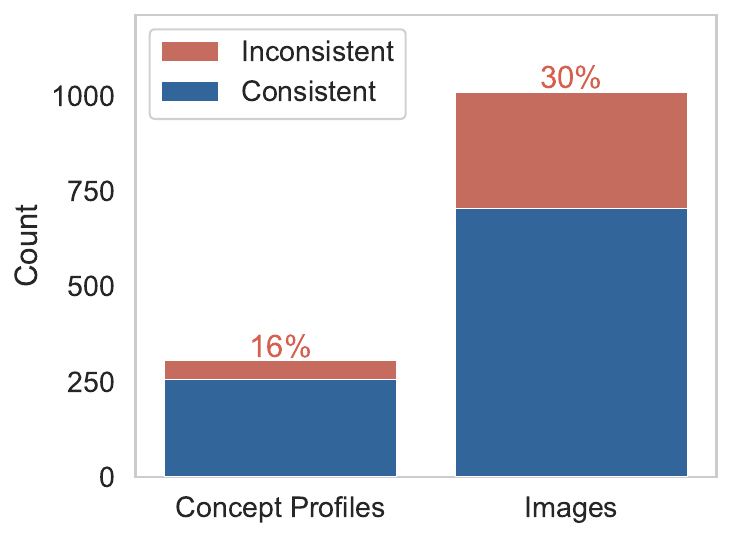}
    \caption{Dataset partition}
    \end{subfigure}
    \begin{subfigure}{0.48\textwidth}
    \centering
    \includegraphics[width=\textwidth]{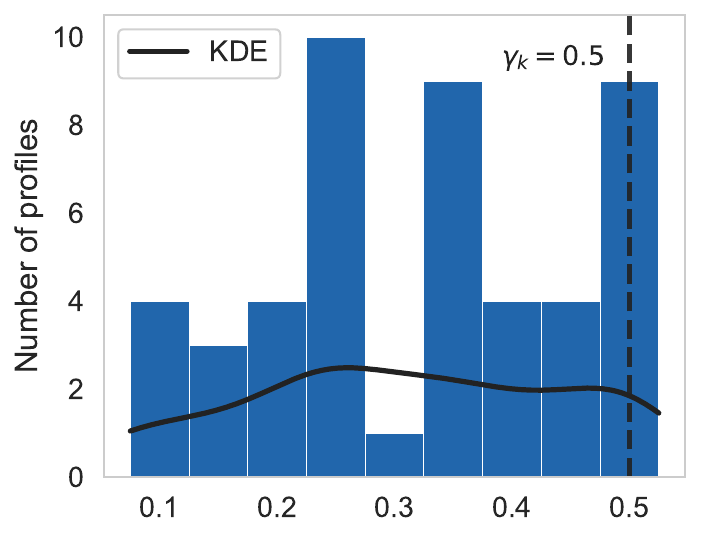}
    \caption{Distribution of $\gamma_k$}
    \end{subfigure}
    
    \caption{Rough set partition of the Derm7pt dataset. (a) Stacked bar chart showing the consistent (blue) and inconsistent (red) split at the concept-profile level and the image level. (b) Histogram of conflict ratio $\gamma_k$ across all 50 inconsistent profiles, with kernel density estimate. The dashed vertical line marks maximum ambiguity $\gamma_k=0.5$.}
    \label{fig:partition}
\end{figure}

Figure~\ref{fig:partition}(a) shows the profile- and image-level partition. The conflict-ratio distribution is shown in Figure~\ref{fig:partition}(b), where we can see a markedly non-uniform distribution. A cluster of profiles near $\gamma_k \approx 0.25$ represents cases where one class predominates with a small number of outlier annotations, while 9 profiles (18.0\% of inconsistent profiles) reach the maximum ambiguity $\gamma_k = 0.5$, meaning the concept signature provides no discriminative signal. The distribution mean ($\bar{\gamma}=0.31$, $s=0.13$) is above the annotation-noise range ($\gamma_k < 0.2$), which would indicate genuine uncertainty.
	
Table~\ref{tab:top5} lists the five most ambiguous profiles observed in the dataset (reaching the highest $\gamma_k$ score possible). These profiles involve irregular dots and globules frequently in combination with a blue-whitish veil or irregular streaks, pairings that are known to overlap clinically between melanoma and pigmented Spitz nevi \cite{massone2021dermoscopic}. The morphological continuum between regular and irregular globules and the visual similarity between melanoma blue-whitish veil and the blue-gray peppering of regressing nevi are well-documented sources of inter-observer disagreement in dermoscopy \cite{rodriguez2022concordance}.

\begin{table}[!ht]
    \centering
    \caption{Five inconsistent concept profiles with the highest conflict ratio $\gamma_k$, where $n_k$ is the total number of images in profile $k$, melanoma count is $n_k^{\mathrm{mel}}$, and non-melanoma count is $n_k^{\mathrm{nmel}}$. Only active concept values are listed.}
    \label{tab:top5}
    \small
    \setlength{\tabcolsep}{4pt}
    \begin{tabular}{@{}cccccl@{}}
        \toprule
        \# & $n_k$ & $n_k^{\mathrm{mel}}$ & $n_k^{\mathrm{nmel}}$ & $\gamma_k$ & Key concept values \\
        \midrule
        1 & 4 & 2 & 2 & 0.5 & Irregular streaks, irregular dots and globules, blue-whitish veil \\
        2 & 2 & 1 & 1 & 0.5 & Diffuse irregular pigmentation, white areas regression \\
        3 & 2 & 1 & 1 & 0.5 & Diffuse irregular pigmentation, regular dots and globules, hairpin vascular \\
        4 & 2 & 1 & 1 & 0.5 & Atypical pigment network, irregular streaks, blue areas regression, regular dots and globules \\
        5 & 2 & 1 & 1 & 0.5 & Atypical pigment network, irregular streaks, localized irregular pigmentation \\
        \bottomrule
    \end{tabular}
\end{table}
	
\begin{figure}[!ht]
    \centering
    \includegraphics[width=\textwidth]{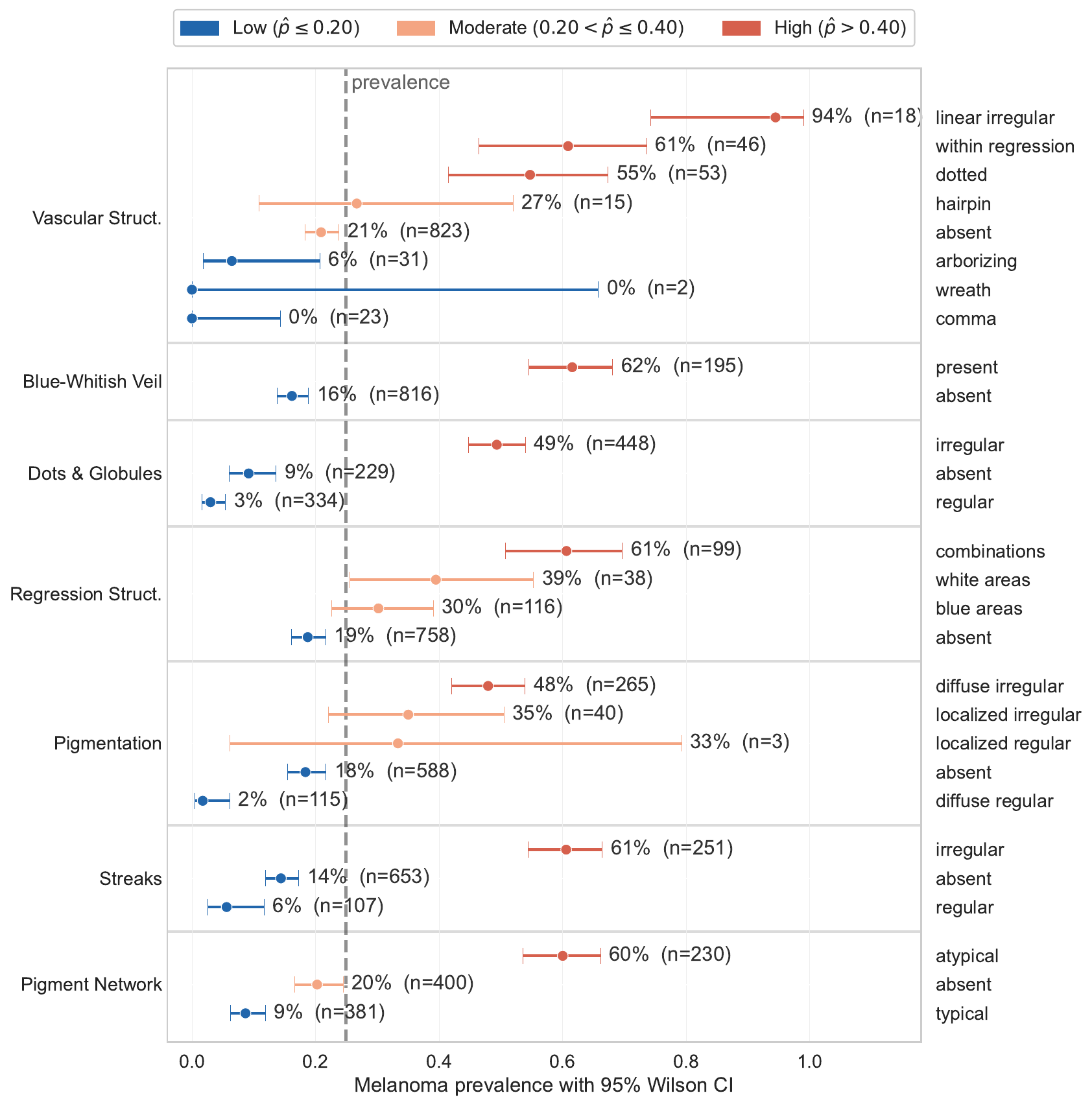}
    \caption{Melanoma prevalence per concept value with 95\% Wilson confidence intervals, where $\hat{p}$ is the estimated melanoma proportion for each concept value. Color encodes risk tier: high ($\hat{p}>0.40$), moderate ($0.20 < \hat{p} \le 0.40$), low ($\hat{p} \le 0.20$). The dotted vertical line marks the dataset-level melanoma prevalence (24.9\%). Right-axis labels group concept values by their parent attribute in the set of dermoscopic concepts $C$.}
    \label{fig:melanoma_rate}
\end{figure}

Figure~\ref{fig:melanoma_rate} shows the melanoma prevalence $\hat{p}$ for each concept value with 95\% Wilson confidence intervals \cite{wilson1927probable}, where $\hat{p}$ denotes the estimated proportion of melanoma cases among all images carrying that concept value. Three concept values stand out at high melanoma rate with sufficient sample sizes: irregular streaks ($\hat{p}=0.61$, $n=251$), blue-whitish veil present ($\hat{p}=0.62$, $n=195$), and atypical pigment network ($\hat{p}=0.60$, $n=230$). The high-risk concept values are enriched in the boundary region, where irregular dots and globules appear in 79\% of inconsistent profile images compared to 29\% of consistent profile images. Irregular streaks appear in 39\% versus 19\%, and blue-whitish veil is present in 29\% versus 15\%.
	
\begin{figure}[!ht]
    \centering
    \begin{subfigure}{0.48\textwidth}
    \centering
    \includegraphics[width=\textwidth]{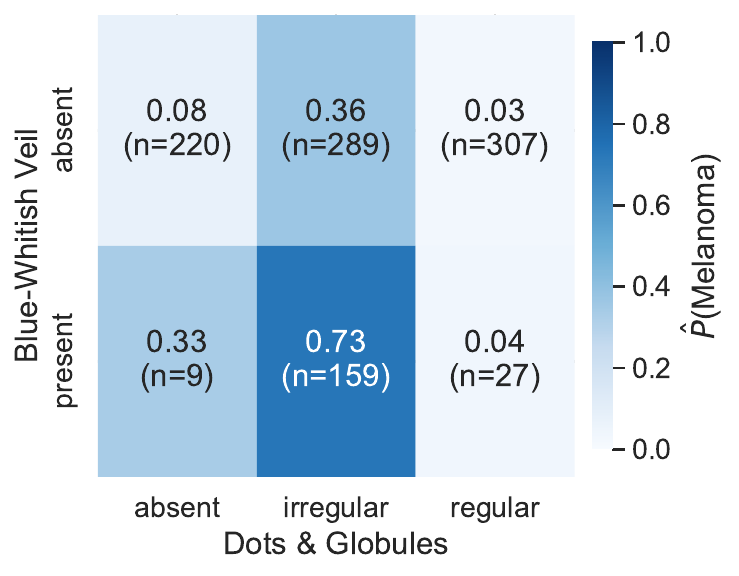}
    \caption{}
    \end{subfigure}
    \begin{subfigure}{0.48\textwidth}
    \centering
    \includegraphics[width=\textwidth]{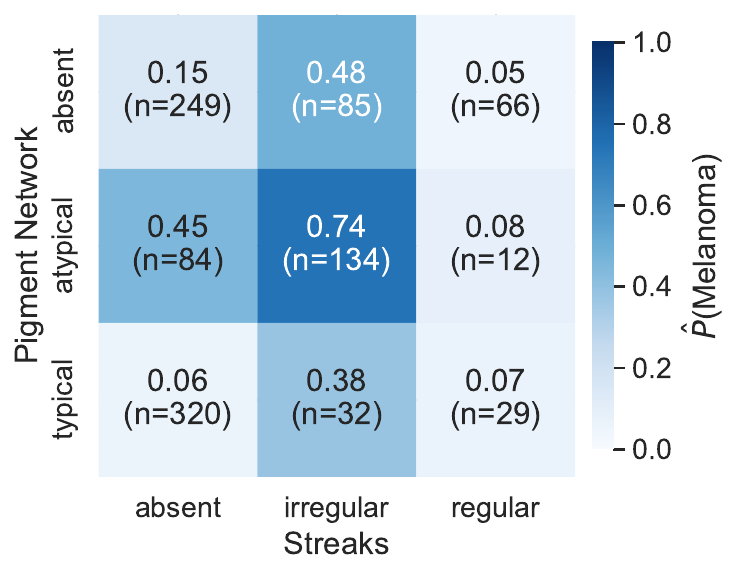}
    \caption{}
    \end{subfigure}
    
    \caption{Joint melanoma rate heatmaps. Each cell shows the estimated probability $\hat{P}(\text{melanoma})$ and the sample count $n$ for images that carry both concept values. Cells with $n<3$ are omitted. (a) Blue-whitish veil $\times$ dots and globules. (b) Pigment network $\times$ streaks. The color scale is shared across both panels.}
    \label{fig:cooccurrence}
\end{figure}

Figure~\ref{fig:cooccurrence} shows joint melanoma rate heatmaps for two high-information concept pairs. In this visualization, each cell reports the estimated joint probability $\hat{P}(\text{melanoma})$ for dermoscopic images carrying both concept values. The combination blue-whitish veil present $\times$ irregular dots and globules yields $\hat{P}(\text{melanoma})=0.73$ ($n=159$). However, the single most frequent boundary-region entry in this heatmap is absent blue-whitish veil $\times$ irregular dots and globules. This scenario accounts for 154 of the 306 boundary images despite a lower joint melanoma probability of 0.36. For the pigment network $\times$ streaks pair, the atypical $\times$ irregular cell reaches $\hat{P}=0.74$ ($n=134$) and is also the most boundary-concentrated cell with 85 images. The fact that these high-risk cells contribute disproportionately to the boundary region $\mathrm{BND}_C(d)$ confirms that inconsistency is concentrated precisely where diagnostic stakes are highest.
	
\begin{figure}[!ht]
    \centering
    
    \begin{subfigure}{0.50\textwidth}
    \centering
    \includegraphics[width=0.9\textwidth]{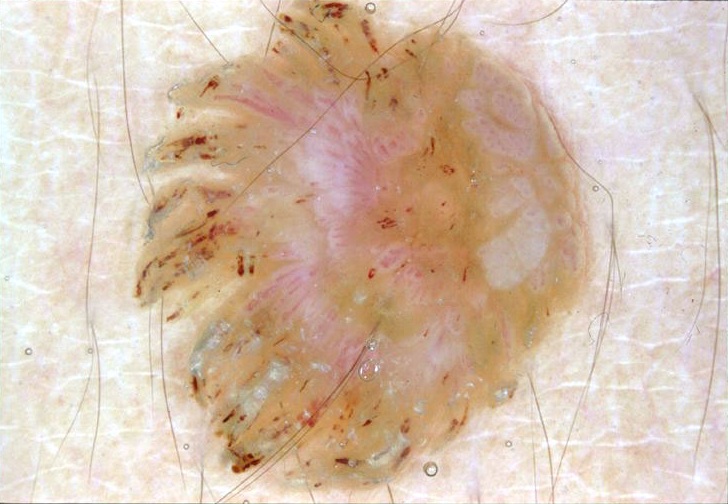}
    \caption{seborrheic keratosis}
    \end{subfigure}
    \begin{subfigure}{0.48\textwidth}
    \centering
    \includegraphics[width=0.885\textwidth]{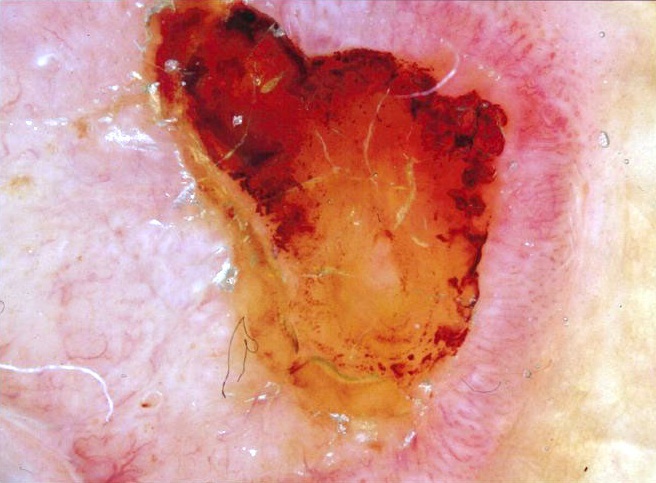}
    \caption{melanoma}
    \end{subfigure}
    
    \caption{Inconsistent pair from profile P1 ($\gamma_1=0.50$). Despite identical concept descriptions, the left image is labeled as seborrheic keratosis ($d=0$), whereas the right image is labeled as melanoma ($d=1$).}
    \label{fig:inconsistency_example}
\end{figure}

Figure~\ref{fig:inconsistency_example} illustrates the inconsistency with a concrete example involving two images. Image \texttt{Ael/Ael498.jpg} is labeled as melanoma, while image \texttt{Ael/Ael427.jpg} is labeled as seborrheic keratosis. Both share the identical concept signature under $C$ where all seven dermoscopic attributes $a_1,\ldots,a_7$ are absent except for hairpin vascular structures. Because these two images are indiscernible under $\mathrm{IND}(C)$, they belong to the same equivalence class $[x]_C$, and no CBM can correctly classify both. Profile P1 has $\gamma_1 = 0.5$ such that the concept signature provides no discriminative signal. In these cases,  the accuracy-ceiling contribution of this profile is determined by a coin flip over two equally represented classes.

Image \texttt{Ael/Ael427.jpg}, diagnosed as seborrheic keratosis, shows dermoscopic traits that align well with classical descriptions of keratinizing lesions. The lesion has a waxy, yellow-brown appearance with a lobulated surface pattern and several peripheral loop vessels surrounded by a pale halo. These vessels correspond to the well-known hairpin morphology frequently reported in seborrheic keratosis. Pigment network structures are not visible, and streaks are absent. Pigmentation is present but confined to localized keratin areas rather than distributed irregularly across the lesion. No regression structures are visible, globules are absent, and the blue-whitish veil is not observed. The lesion also appears dome-shaped and elevated relative to the surrounding skin. A concept vector that matches these visual characteristics could therefore be written as \textit{pigment network}=absent, \textit{streaks}=absent, \textit{pigmentation}=localized regular, \textit{regression structures}=absent, \textit{dots and globules}=absent, \textit{blue whitish veil}=absent, \textit{vascular structures}=hairpin. Yet, this must be validated by dermatologists.

In contrast, image \texttt{Ael/Ael498.jpg}, labeled melanoma, presents a markedly different dermoscopic morphology. The lesion displays strong asymmetry, extensive red coloration, and multiple irregular hemorrhagic areas. Pigment network structures are not visible, and streaks are absent. Pigmentation is clearly present and distributed in an irregular, diffuse manner rather than being absent. Regression structures and globules are not evident, and a blue-whitish veil is not visible in the photograph. The vascular morphology does not correspond to symmetric loop hairpin vessels. Instead, the visible vessels appear fragmented and irregular, a pattern compatible with linear irregular vessels frequently associated with melanoma. A concept vector that reflects these observations can therefore be written as \textit{pigment network}=absent, \textit{streaks}=absent, \textit{pigmentation}=diffuse irregular, \textit{regression structures}=absent, \textit{dots and globules}=absent, \textit{blue-whitish veil}=absent, \textit{vascular structures}=linear irregular. As in the previous case, such an annotation should be approved by experts.

Dataset inconsistency undermines the semantic validity of concept attributions. The CBM requires that the composition $h \circ \hat{\mathbf{v}}: U \to \{0,1\}$ be well-defined, where $\hat{\mathbf{v}}(x) \in \{0,1\}^7$ is the predicted concept vector (the values predicted by the model for each attribute in $C$) and $h$ maps concept vectors to a diagnosis. Inconsistency violates this requirement since there exist pairs $x_i, x_j \in \mathrm{BND}_C(d)$ such that $\hat{\mathbf{v}}(x_i) = \hat{\mathbf{v}}(x_j)$ but $d(x_i) \ne d(x_j)$, making $h$ undefined on those inputs at the concept level. When the model must simultaneously predict both labels from the same concept vector, gradient-based attribution methods will attribute the difference to feature correlations that bypass the bottleneck, producing explanations that capture artifacts rather than clinical insight and silently violating the core CBM interpretability guarantee.

\section{Adressing Inconsistency}

We acknowledge that inconsistencies in concept annotations are an inherent reality of real-world medical datasets. It often arises from inter-observer variability, ambiguous lesion morphology, and the inherent subjectivity of dermoscopic interpretation. However, boundary region images should not be used as training data for CBMs, since exposing the model to contradictory supervision undermines the core guarantee that the concept bottleneck functions as a well-defined mapping. The appropriate choice is to train on clean, consistent data and allow the model to generalize to ambiguous real-world cases at inference time. Several strategies exist to address concept inconsistency at the dataset level. Reannotation by domain experts is the most solid option, but it is costly, requires specialized dermoscopic knowledge, and risks introducing new disagreements. Expanding the concept set could break existing inconsistencies by making previously indistinguishable profiles separable, but it requires the new annotations to be collected exhaustively across the entire dataset. A third and more practical direction is to remove the inconsistent instances from the dataset entirely. In this section, we explore two strategies that follow this direction, both yielding a cleaner version of the benchmark that we refer to as Derm7pt+.
    
\subsection{Filtering Strategies}
	
The existence of $\mathrm{BND}_C(d)$ means that any CBM trained on the raw dataset will be exposed to contradictory supervision where the same concept prediction vector $\hat{\mathbf{v}}$ is paired with both $d=1$ and $d=0$ labels. In this subsection, we will discuss two filtering strategies rooted in the presumably correct reasoning \cite{napoles2023pids} to remove inconsistent samples from the dataset. It should be highlighted that we do not seek to remove difficult instances from the dataset, but those that are impossible to classify by a CBM that determines the class labels solely from the concept heads.
	
The first strategy concerns symmetric removal, which restricts the dataset to $\mathrm{POS}_C(d)$, removing all samples in $\mathrm{BND}_C(d)$ regardless of label. This can be formalized as follows:
\begin{equation}
    U'
    = \bigl\{x \in U : [x]_C \subseteq d^{-1}(0)\bigr\}
    \cup
    \bigl\{x \in U : [x]_C \subseteq d^{-1}(1)\bigr\}.
    \label{eq:dsym}
\end{equation}
	
Symmetric filtering gives $|U'| = 705$ ($-30.3\%$), and by construction $\gamma(C,d)|_{U'} = 1.0$. Hence, every concept profile maps to a unique, unambiguous label, which fully restores the CBM guarantee that $h \circ \hat{\mathbf{v}}$ is a valid function. The cost is the removal of 136 melanoma images (54.0\% of all melanoma), worsening the class imbalance from 1:3.0 to 1:5.1.
	
Asymmetric removal retains melanoma-labeled images in inconsistent profiles, removing only the conflicting non-melanoma samples. This embeds the clinical prior that false negatives carry a higher cost than false positives in a screening setting. It can be seen as an undersampling method and can be formalized as follows:
\begin{equation}
    U'' = U' \cup \bigl\{x \in \mathrm{BND}_C(d) : d(x) = 1\bigr\}.
    \label{eq:dasym}
\end{equation}
	
\begin{figure}[!ht]
    \centering
    \begin{subfigure}{0.48\textwidth}
    \centering
    \includegraphics[width=\textwidth]{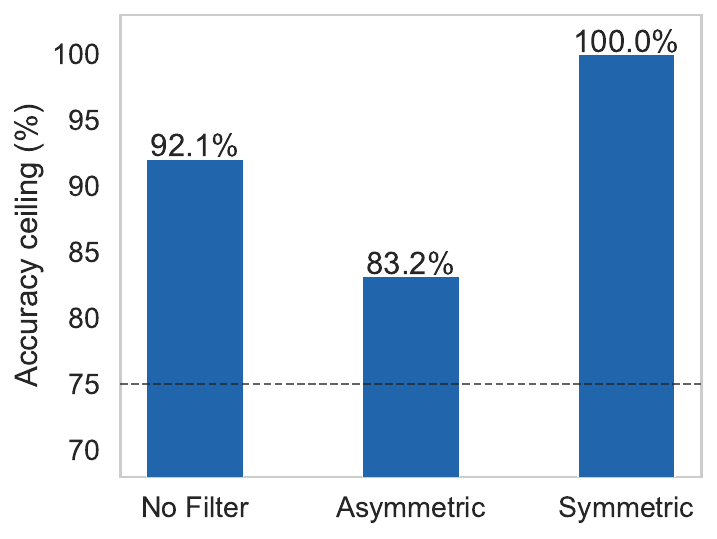}
    \caption{Theoretical accuracy ceiling}
    \end{subfigure}
    \begin{subfigure}{0.49\textwidth}
    \centering
    \includegraphics[width=\textwidth]{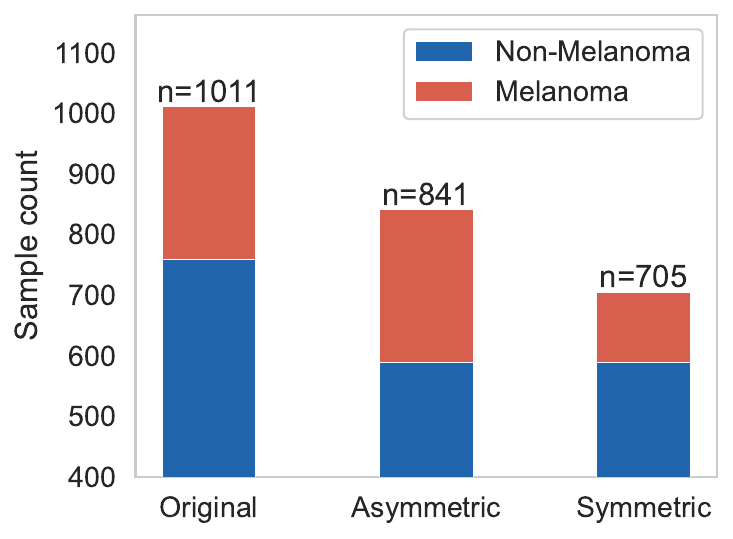}
    \caption{Post-filter dataset composition}
    \end{subfigure}
    \caption{Filtering strategy comparison. (a) Theoretical accuracy ceiling for each configuration. The dashed line marks the majority-class baseline (75.1\%). (b) Post-filter dataset composition by class.}
    \label{fig:strategies}
\end{figure}

The asymmetric filtering approach gives $|U''| = 841$ ($-16.8\%$). All 252 melanoma images are retained, and the class ratio improves slightly to 1:2.3. The dataset remains partially inconsistent, but the decision boundary shifts in the clinically conservative direction. The hard accuracy ceiling on $U''$ rises to 83.2\%. Table~\ref{tab:strategies} summarizes the trade-offs, while Figure~\ref{fig:strategies} visualizes the ceilings and post-filter compositions.
	
\begin{table}[!ht]
    \centering
    \caption{Comparison of filtering strategies such that $\gamma(C,d)$ is the quality of classification defined in Equation~(\ref{eq:gamma}), $\mathrm{acc}^*$ is the accuracy ceiling defined in Equation~(\ref{eq:ceil}), and M:NM is the melanoma to non-melanoma class ratio.}
    \label{tab:strategies}
    \small
    \setlength{\tabcolsep}{5pt}
    \begin{tabular}{@{}lcccc@{}}
        \toprule
        Property & No filter & Asymmetric & Symmetric \\
        \midrule
        Number of dermoscopic images                   & 1,011 & 841      & 705   \\
        Number of melanoma samples retained            & 252   & 252      & 116   \\
        Class imbalance ratio (M:NM)                   & 1:3.0 & 1:2.3    & 1:5.1 \\
        Quality of classification $\gamma(C,d)$        & 0.697 & $>$0.697 & 1.000 \\
        Accuracy ceiling score $\mathrm{acc}^*$        & 92.1\% & 83.2\%  & 100\% \\
        All melanoma samples preserved                 & Yes   & Yes      & No    \\
        Concept consistency assessment                 & No    & Partial  & Full  \\
        \bottomrule
    \end{tabular}
\end{table}
	
For rigorous interpretability benchmarking, symmetric removal ($U'$) is preferred since it yields a fully consistent dataset with $\gamma = 1$ and a well-defined accuracy ceiling of 100\%. For clinical screening applications, asymmetric removal ($U''$) is preferred since it retains all melanoma training signal, preserves sensitivity, and imposes only bounded residual inconsistency. Neither strategy is universally optimal since both come at the cost of losing valuable data, so the choice should be documented explicitly in any publication using Derm7pt+ for CBM evaluation.

Beyond filtering the concept annotations to remove inconsistent patterns, we manually cropped all images to remove border artifacts and better center the lesion within the frame. All images were saved at 384×384 pixels. In principle, a CNN backbone trained with enough samples and epochs would learn to distinguish diagnostically relevant regions from noise at the image borders. However, both filtering strategies substantially reduce the number of available training samples, implying that longer training schedules risk overfitting rather than improving generalization. This motivated our choice of a conservative training setup, as described in the next section.

\section{Hard Concept Bottleneck Model}

We now build and evaluate a hard CBM to assess how well concept-based classifiers perform on Derm7pt+ under both filtering strategies. The original unfiltered Derm7pt dataset is not considered in this evaluation since it has been sufficiently explored in the literature \cite{hou2024caw, lucieri2022exaid, bie2024mica}. Moreover, as the analysis in Section \ref{sec:analysis} makes clear, training a CBM on concept-inconsistent data exposes the model to contradictory signals that undermine the core guarantee of the bottleneck.

\subsection{Data Preparation}

All experiments use the symmetric filtering strategy defined in Equation~(\ref{eq:dsym}), which restricts the dataset to the positive region $\mathrm{POS}_C(d)$ by removing every image whose concept profile belongs to $\mathrm{BND}_C(d)$. As reported in Table~\ref{tab:strategies}, this yields $|U'| = 705$ images with a quality of classification $\gamma(C,d) = 1.0$, guaranteeing that every concept profile maps unambiguously to a unique diagnosis label. The theoretical accuracy ceiling on $U'$ is therefore 100\%, removing the hard constraint that would otherwise limit any concept-only classifier trained on the raw dataset.

The Derm7pt benchmark provides predefined training, validation, and test splits. The symmetric filtering defined in Equation~(\ref{eq:dsym}) is applied to each split, removing every image whose concept profile belongs to $\mathrm{BND}_C(d)$ as identified from the full dataset. The official partition is otherwise preserved exactly, so no resplitting or pooling is performed.

Training images undergo a stochastic augmentation pipeline comprising random resized cropping (scale $\in [0.75, 1.0]$, output size $224 \times 224$), horizontal and vertical flips, random rotation up to $90^{\circ}$, random affine transforms (translation up to 10\%, shear up to $15^{\circ}$, applied with probability 0.5), color jitter (brightness 0.4, contrast 0.4, saturation 0.3, hue 0.1), random grayscale conversion (probability 0.05), and a randomly selected sharpening, Gaussian blur, or identity transform. Validation and test images are resized to $224 \times 224$ and normalized only, using ImageNet statistics $\mu = (0.485, 0.456, 0.406)$ and $\sigma = (0.229, 0.224, 0.225)$. Class imbalance in the label space is handled through balanced class weights computed on the training set via the inverse-frequency formula depicted below:
\begin{equation}
    w_\ell = \frac{|P_{\mathrm{train}}|}{2 \cdot |d^{-1}(\ell) \cap P_{\mathrm{train}}|},
    \qquad \ell \in \{0,1\},
    \label{eq:classweight}
\end{equation}

Analogously, per-class weights are computed independently for each concept attribute in $C$ to address concept-level imbalance. Such a strategy is necessary since some concept states are underspecified in the annotations.

\subsection{Hard CBM Architecture}

The model follows the hard Concept Bottleneck architecture introduced by \cite{koh2020cbm}. It comprises three components: (i) a backbone feature extractor $\phi_\theta$, (ii) a set of concept prediction heads $\{g_c\}_{c \in C}$, and (iii) a final label classifier $h_\psi$. Given an input image $x \in U$, the backbone produces a feature vector $\mathbf{z} = \mathrm{pool}(\phi_\theta(x)) \in \mathbb{R}^D$ via global average pooling. For each concept $c \in C$, a linear head maps $\mathbf{z}$ to a vector of logits over the concept categories:
\begin{equation}
    \mathbf{l}_c = g_c(\mathbf{z}) = W_c\,\mathbf{z} + b_c \in \mathbb{R}^{|V_c|},
    \label{eq:concepthead}
\end{equation}
where $V_c$ represents the set of possible values for concept $c$. The bottleneck representation is obtained by converting each logit vector into a hard one-hot encoding, as shown below:
\begin{equation}
    \hat{\mathbf{v}}_c = \mathrm{one\text{-}hot}\!\left(\arg\max_j\, [\mathbf{l}_c]_j\right) \in \{0,1\}^{|V_c|}.
    \label{eq:onehot}
\end{equation}

The full concept bottleneck vector is the concatenation $\hat{\mathbf{v}} = [\hat{\mathbf{v}}_{c_1}, \ldots, \hat{\mathbf{v}}_{c_7}] \in \{0,1\}^M$, where $M = \sum_{c \in C} |V_c|$. This hard binarization ensures that the label classifier receives only the discrete concept assignments that a clinician would observe. The label classifier $h_\psi$ is a two-layer network applied to $\hat{\mathbf{v}}$, composed of a linear layer of hidden dimension 128 followed by layer normalization, ReLU activation, a dropout layer, and a final linear layer.

\begin{figure}[!ht]
    \centering
    \includegraphics[width=0.8\textwidth]{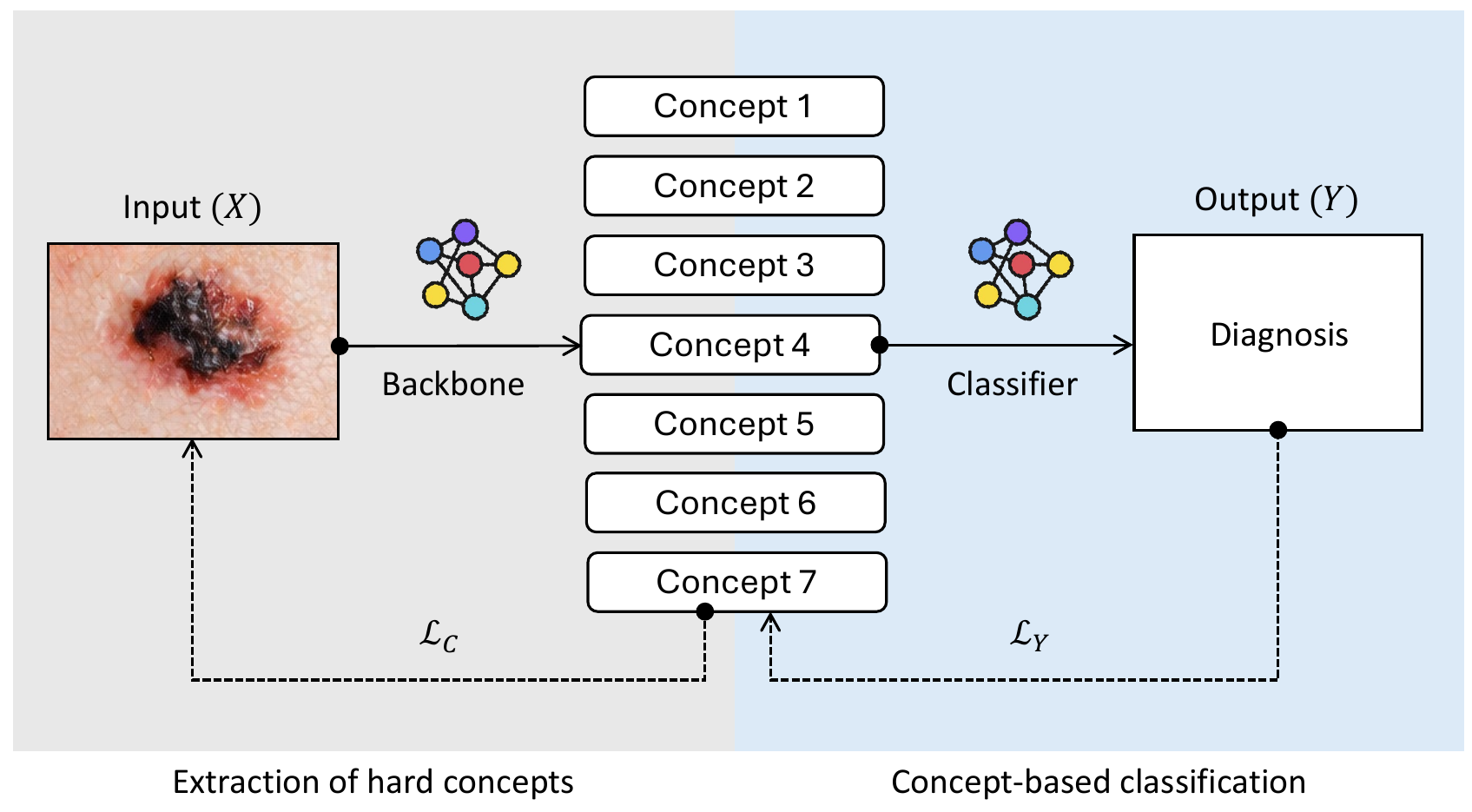}
    \caption{Architecture of our hard CBM with stop-gradient. An input image $x$ is passed through a backbone $\phi_{\theta}$ to produce a feature vector $\mathbf{z}$. Separate concept heads $\{g_c\}$ predict logits for each dermoscopic concept. These logits are converted into a single hard, discrete concept vector $\hat{\mathbf{v}}$ via argmax and one-hot encoding. To prevent label leakage, the gradient flow is detached at the bottleneck, ensuring the label classifier $h_{\psi}$ and its loss $L_Y$ do not influence the feature extractor. The backbone and concept heads are updated solely by the concept loss $L_C$, preserving the semantic purity of the bottleneck.}
    \label{fig:bottleneck}
\end{figure}

The backbone $\phi_\theta$, concept heads $\{g_c\}$, and label classifier $h_\psi$ are trained jointly in a single stage, avoiding the sequential two-stage procedure of the original CBM formulation. However, joint training introduces the risk of label leakage where gradient signals from the label loss could propagate back through $\hat{\mathbf{v}}$ and distort the concept representations. To prevent this, the gradient flow through the hard bottleneck is severed by treating $\hat{\mathbf{v}}$ as a stop-gradient tensor during the backward pass, $\hat{\mathbf{v}}_{\mathrm{sg}} = \mathrm{sg}(\hat{\mathbf{v}})$, so that the label classifier loss does not reach the backbone or the concept heads. As a result, $\phi_\theta$ and $\{g_c\}$ are updated exclusively by the concept supervision signal, preserving the semantic alignment of the bottleneck without sacrificing training efficiency. Figure \ref{fig:bottleneck} shows the blueprint of the hard CBM explored in this paper.

The total training loss combines a concept prediction term and a label prediction term:
\begin{equation}
    L = \frac{1}{N}\sum_{i=1}^N \bigl[
        \lambda_1\,L_C(x_i, \mathbf{v}_i)
        + \lambda_2\,L_Y(x_i, y_i)
    \bigr],
    \label{eq:totalloss}
\end{equation}
where $\lambda_1, \lambda_2 \geq 0$ are balancing coefficients. The concept loss is the mean of per-concept weighted cross-entropies:
\begin{equation}
    L_C(x_i, \mathbf{v}_i) = \frac{1}{|C|}\sum_{c \in C}
        \mathrm{CE}\!\left(\mathbf{l}_c^{(i)},\, v_c^{(i)},\, \mathbf{w}_c\right),
    \label{eq:conceptloss}
\end{equation}
where $v_c^{(i)} \in V_c$ is the ground-truth value for concept $c$ and $\mathbf{w}_c \in \mathbb{R}^{|V_c|}$ is the vector of class weights computed via Equation~(\ref{eq:classweight}) for concept $c$. The label loss is a weighted cross-entropy:
\begin{equation}
    L_Y(x_i, y_i) = \mathrm{CE}\!\left(\hat{y}^{(i)},\, y_i,\, \mathbf{w}_y\right),
    \label{eq:labelloss}
\end{equation}
with $\mathbf{w}_y = (w_0, w_1)$ computed from Equation~(\ref{eq:classweight}).

\subsection{Empirical Results}

Nineteen backbone architectures are evaluated: EfficientNet-B0 through B7 (8 models) \cite{Tan2019Efficient}, DenseNet-121, 161, 169, and 201 (4 models) \cite{DenseNet2017}, ResNet-18, 34, 50, 101, and 152 (5 models) \cite{He_2016_CVPR}, and Wide ResNet-50-2 and 101-2 (2 models)\cite{Zagoruyko2016wide}. All backbones are initialized with ImageNet pretrained weights. The fraction of tunable backbone layers, the batch size, and the initial learning rate were treated as hyperparameters and selected by grid search on the validation set over the ranges $\{50\%, 75\%, 100\%\}$, $\{16, 32\}$, and $\{10^{-3}, 5\times10^{-4}, 10^{-4}\}$, respectively. The remaining hyperparameters were fixed after preliminary experimentation, resulting in weight decay $10^{-4}$ for the backbone, dropout rate 0.3, hidden dimension 128 for the label classifier, cosine annealing with $\eta_{\min} = 10^{-6}$, and loss balancing coefficients $\lambda_1 = \lambda_2 = 1.0$. All models are trained for up to 20 epochs since we observed that longer training leads to overfitting, even using early stopping. The best checkpoint, defined as the one that achieves the highest macro-averaged F1 on the validation label set, is used for test set evaluation.

Label performance is measured by both accuracy and macro-averaged F1. F1 is the primary evaluation criterion because the test set remains class-imbalanced after filtering, and accuracy can be misleading under imbalance. For concept prediction, accuracy is the primary metric, as the concept prediction task does not require every concept to be correctly identified simultaneously for the final label to be correct. Furthermore, the concept annotations in Derm7pt carry intrinsic noise from inter-observer variability in dermoscopic attribute assignment, so macro-averaged concept accuracy captures overall correctness across attributes without penalizing modest disagreement on noisy categories as heavily as F1 would.

Results are reported separately for the symmetric and asymmetric filtering strategies. The symmetric test set contains only concept-consistent images, while the asymmetric test set also includes melanoma images from the boundary region. Note that performance scores across strategies are not directly comparable since they lead to different datasets.

Table~\ref{tab:backbone_sym} reports performance under the symmetric strategy. Label accuracy ranges from 0.74 (DenseNet-121) to 0.90 (EfficientNet-B5), and label F1 from 0.67 (DenseNet-121) to 0.85 (EfficientNet-B5). Concept accuracy ranges from 0.45 (DenseNet-121) to 0.70 (EfficientNet-B5 and others), and concept F1 from 0.39 (DenseNet-121) to 0.56 (EfficientNet-B5 and others). The EfficientNet family achieves the highest label F1 scores across all four families, with EfficientNet-B5 attaining the best performance overall. DenseNet models lag in label F1, with the best result (DenseNet-169, F1~=~0.78) falling seven points below EfficientNet-B5. ResNet-152 achieves the best label F1 within the ResNet family (0.81), while ResNet-34 is the weakest across all architectures in both label accuracy (0.80) and label F1 (0.71). It can be concluded that Wide ResNet models offer no advantage over their standard-width counterparts on this task.

\begin{table}[!ht]
    \centering
    \caption{Backbone comparison on the Derm7pt test set under symmetric filtering (concept-consistent images only). The best checkpoint by validation label F1 is used for evaluation. Bold values indicate the best results overall.}
    \label{tab:backbone_sym}
    \small
    \setlength{\tabcolsep}{5pt}
    \begin{tabular}{@{}clccccc@{}}
        \toprule
        Family & Backbone & Best Epoch & Label Acc. & Label F1 & Concept Acc. & Concept F1 \\
        \midrule
        \multirow{8}{*}{\textit{EfficientNet}}
        & efficientnet\_b0 & 19 & 0.85 & 0.77 & 0.70 & 0.56 \\
        & efficientnet\_b1 & 14 & 0.88 & 0.82 & 0.70          & 0.56          \\
        & efficientnet\_b2 & 10 & 0.85 & 0.77 & 0.66          & 0.53          \\
        & efficientnet\_b3 & 11 & 0.89 & 0.82 & 0.67          & 0.53          \\
        & efficientnet\_b4 & 12 & 0.89 & 0.78 & 0.70          & 0.53          \\
        & efficientnet\_b5 & 13 & \textbf{0.90} & \textbf{0.85} & \textbf{0.70} & \textbf{0.56} \\
        & efficientnet\_b6 &  8 & 0.89 & 0.79 & 0.65          & 0.50          \\
        & efficientnet\_b7 & 20 & 0.87 & 0.80 & 0.69          & 0.54          \\
        \midrule
        \multirow{4}{*}{\textit{DenseNet}}
        & densenet121      &  4 & 0.74 & 0.67 & 0.45 & 0.39 \\
        & densenet161      & 16 & 0.86 & 0.77 & 0.62 & 0.50 \\
        & densenet169      & 15 & 0.86 & 0.78 & 0.64 & 0.51 \\
        & densenet201      & 12 & 0.86 & 0.77 & 0.64 & 0.50 \\
        \midrule
        \multirow{5}{*}{\textit{ResNet}}
        & resnet18         &  9 & 0.84 & 0.74 & 0.63 & 0.49 \\
        & resnet34         & 16 & 0.80 & 0.71 & 0.57 & 0.46 \\
        & resnet50         & 11 & 0.80 & 0.73 & 0.61 & 0.49 \\
        & resnet101        & 14 & 0.88 & 0.80 & 0.69 & 0.53 \\
        & resnet152        & 18 & 0.89 & 0.81 & 0.69 & 0.54 \\
        \midrule
        \multirow{2}{*}{\textit{Wide ResNet}}
        & wide\_resnet50\_2  & 16 & 0.87 & 0.75 & 0.70 & 0.53 \\
        & wide\_resnet101\_2 & 10 & 0.84 & 0.72 & 0.66 & 0.49 \\
        \bottomrule
    \end{tabular}
\end{table}

The observed spread between the best and worst concept accuracy scores (0.70 versus 0.45) suggests that concept prediction quality is less sensitive to backbone choice than label prediction quality. This pattern aligns with the noisy nature of Derm7pt concept annotations, where inter-observer disagreements impose a floor on concept accuracy that is not subject to the backbone choice. However, such a precision level is not required for accurate label approximation.

Table~\ref{tab:backbone_asym} reports performance under the asymmetric strategy, which retains all melanoma images from the boundary region and removes only the conflicting non-melanoma samples. Label accuracy ranges from 0.72 (ResNet-18) to 0.85 (EfficientNet-B1 and B7), and label F1 from 0.68 (ResNet-18) to 0.82 (EfficientNet-B7). Concept accuracy ranges from 0.55 (DenseNet-161) to 0.70 (EfficientNet-B7), and concept F1 from 0.43 (DenseNet-161) to 0.56 (EfficientNet-B7). EfficientNet-B7 achieves the best results across all four metrics under asymmetric filtering. DenseNet-161 is the weakest architecture under this strategy across all metrics, a reversal from its relatively competitive concept F1 under symmetric filtering. ResNet-18, which performed reasonably well under symmetric filtering, drops sharply in label F1 (0.68) under asymmetric filtering, which suggests sensitivity to the residual inconsistency introduced by the retained boundary melanoma images.

\begin{table}[!ht]
    \centering
    \caption{Backbone comparison on the Derm7pt test set under asymmetric filtering (all melanoma images retained). The best checkpoint by validation label F1 is used for evaluation. Bold values indicate the best results overall.}
    \label{tab:backbone_asym}
    \small
    \setlength{\tabcolsep}{5pt}
    \begin{tabular}{@{}clccccc@{}}
        \toprule
        Family & Backbone & Best Epoch & Label Acc. & Label F1 & Concept Acc. & Concept F1 \\
        \midrule
        \multirow{8}{*}{\textit{EfficientNet}}
        & efficientnet\_b0 &  7 & 0.80 & 0.77 & 0.67          & 0.54          \\
        & efficientnet\_b1 & 12 & 0.85 & 0.81 & 0.69 & 0.56          \\
        & efficientnet\_b2 &  9 & 0.83 & 0.82 & 0.66          & 0.54          \\
        & efficientnet\_b3 & 11 & 0.81 & 0.79 & 0.66          & 0.53          \\
        & efficientnet\_b4 & 15 & 0.84 & 0.81 & 0.68          & 0.54          \\
        & efficientnet\_b5 & 12 & 0.83 & 0.80 & 0.68          & 0.54          \\
        & efficientnet\_b6 & 12 & 0.83 & 0.80 & 0.69          & 0.55          \\
        & efficientnet\_b7 & 18 & \textbf{0.85} & \textbf{0.82} & \textbf{0.70} & \textbf{0.56} \\
        \midrule
        \multirow{4}{*}{\textit{DenseNet}}
        & densenet121      & 14 & 0.80 & 0.75 & 0.65 & 0.52 \\
        & densenet161      &  7 & 0.77 & 0.73 & 0.55 & 0.43 \\
        & densenet169      & 19 & 0.81 & 0.77 & 0.69 & 0.55 \\
        & densenet201      & 16 & 0.83 & 0.81 & 0.66 & 0.55 \\
        \midrule
        \multirow{5}{*}{\textit{ResNet}}
        & resnet18         &  5 & 0.72 & 0.68 & 0.55 & 0.45 \\
        & resnet34         & 20 & 0.77 & 0.72 & 0.61 & 0.49 \\
        & resnet50         & 17 & 0.81 & 0.75 & 0.70 & 0.55 \\
        & resnet101        & 13 & 0.81 & 0.76 & 0.67 & 0.53 \\
        & resnet152        & 19 & 0.82 & 0.77 & 0.70 & 0.54 \\
        \midrule
        \multirow{2}{*}{\textit{Wide ResNet}}
        & wide\_resnet50\_2  & 13 & 0.81 & 0.76 & 0.67 & 0.53 \\
        & wide\_resnet101\_2 & 20 & 0.81 & 0.77 & 0.68 & 0.54 \\
        \bottomrule
    \end{tabular}
\end{table}

The concept accuracy spread under asymmetric filtering (0.70 versus 0.55) is narrower than under symmetric filtering once DenseNet-121 is excluded, and the best concept accuracy values are similar across both strategies. This reinforces the view that the bottleneck quality relies on concept annotation rather than backbone capacity. However, it must be highlighted that we do not need perfect concept accuracy to achieve accurate diagnosis. 

\section{Concluding Remarks}

We presented Derm7pt+, a concept-consistent dermoscopy benchmark derived from Derm7pt through rough set analysis and symmetric filtering of all boundary-region images. The dataset guarantees perfect quality of classification, meaning every concept profile maps unambiguously to a single diagnosis label, removing the hard accuracy ceiling that afflicts any classifier operating exclusively on hard concepts. Alongside the dataset, we presented a hard CBM architecture that trains the backbone, concept heads, and label classifier jointly while severing gradient flow through the hard bottleneck via a stop-gradient operation, which prevents label leakage into the concept representations.

The inconsistency analysis revealed that \textit{irregular dots and globules} and \textit{irregular streaks} are highly responsible for boundary ambiguity, appearing in 79\% and 39\% of inconsistent-profile images, respectively, compared to 29\% and 19\% in consistent profiles. The most ambiguous profiles combine these attributes with a \textit{blue-whitish veil} or \textit{atypical pigment network}, producing concept signatures with no discriminative signal. At the other end of the spectrum,\textit{irregular streaks}, a present \textit{blue-whitish veil}, and \textit{atypical pigment network} are the strongest melanoma indicators, each carrying a melanoma prevalence above 60\%. \textit{Irregular dots and globules} without a \textit{blue-whitish veil} are often found in the boundary region, contributing 154 of the 306 boundary images. These findings confirm that inconsistency is concentrated where diagnostic stakes are highest. Among the 19 backbones evaluated under symmetric filtering, EfficientNet-B5 achieved the best results overall, with a label F1 score of 0.85, a label accuracy of 0.90, a concept accuracy of 0.70, and a concept F1 of 0.56. Under asymmetric filtering, EfficientNet-B7 led across all four metrics with a label accuracy of 0.85, a label F1 of 0.82, a concept accuracy of 0.70, and a concept F1 of 0.56. The modest spread across architectures in both settings suggests that concept annotation noise, rather than backbone choice, is the binding constraint on bottleneck quality.

The primary limitation of this work is the loss of training data imposed by symmetric filtering, which removes 306 images and eliminates 54.0\% of melanoma cases, worsening class imbalance from 1:3.0 to 1:5.1. Asymmetric filtering mitigates this cost by retaining all melanoma images but introduces residual inconsistency. Therefore, neither strategy is universally optimal for all evaluation settings. Future work should investigate soft and probabilistic concept representations that can accommodate annotation uncertainty without discarding data. A further open direction concerns (i) the interpretability of the learned concept bottleneck, which includes the clinical validity of concept attributions, (ii) their alignment with dermatologist reasoning, and (iii) their utility for actionable explanations in screening settings. These aspects remain largely unexplored and constitute a natural extension of the framework proposed in this paper.

\section*{Acknowledgement}
Y. Salgueiro would like to acknowledge the support provided by ANID Fondecyt Regular 1240293 and National Center for Artificial Intelligence CENIA FB210017, Basal ANID.

\bibliography{references}
\end{document}